\documentclass[lettersize,journal]{IEEEtran}
\usepackage{amsmath,amsfonts}
\usepackage{array}
\usepackage[caption=false,font=normalsize,labelfont=sf,textfont=sf]{subfig}
\usepackage{textcomp}
\usepackage{stfloats}
\usepackage{url}
\usepackage{verbatim}
\usepackage{graphicx}
\usepackage{cite}
\usepackage{amsmath}
\usepackage{amsfonts}       
\usepackage{algorithm}
\usepackage{algpseudocode}
\usepackage{amsthm}
\usepackage{extpfeil}
\usepackage{graphicx} 
\usepackage{array}    
\usepackage{caption}  
\usepackage{adjustbox}
\usepackage{subcaption}
\usepackage{float}
\usepackage{tikz}
\usepackage{multirow}
\usepackage[dvipsnames]{xcolor}         
\usepackage{booktabs}
\usepackage{orcidlink}
\usepackage[numbers]{natbib} 
\hypersetup{
colorlinks,
allcolors=blue,
}
\usetikzlibrary{positioning, arrows.meta}
\newtheorem{theorem}{Theorem}
\newtheorem{definition}{Definition}

\begin{document}

\title{PC-Diffusion: Aligning Diffusion Models with Human Preferences via Preference Classifier}

\author{
Shaomeng Wang\textsuperscript{\orcidlink{0009-0005-2350-1099}},
He Wang\textsuperscript{\orcidlink{0009-0009-2077-7929}}, 
Xiaolu Wei\textsuperscript{\orcidlink{0009-0008-5329-6016}}, 
Longquan Dai\textsuperscript{\orcidlink{0000-0001-7652-5135}},
Jinhui Tang\textsuperscript{\orcidlink{0000-0001-9008-222X}},~\IEEEmembership{Senior Member,~IEEE}
\thanks{Shaomeng Wang, He Wang, Xiaolu Wei, Longquan Dai and Jinhui Tang are with the School of Computer Science and Engineering, 
Nanjing University of Science and Technology, Nanjing 210094, China. 
E-mail: \{smw, wanghe, 124106010727, dailongquan, jinhuitang\}@njust.edu.cn. 
Longquan Dai is the corresponding author.},%
}%




\maketitle
\begin{abstract}
Diffusion models have achieved remarkable success in conditional image generation, yet their outputs often remain misaligned with human preferences. To address this, recent work has applied Direct Preference Optimization (DPO) to diffusion models, yielding significant improvements.~However, DPO-like methods exhibit two key limitations:~\textbf{1) High computational cost},~due to the entire model fine-tuning;~\textbf{2)~Sensitivity to reference model quality}, due to its tendency to introduce instability and bias.~To overcome these limitations, we propose a novel framework for human preference alignment in diffusion models (PC-Diffusion), using a lightweight, trainable Preference Classifier that directly models the relative preference between samples. By restricting preference learning to this classifier, PC-Diffusion decouples preference alignment from the generative model, eliminating the need for entire model fine-tuning and reference model reliance.~We further provide theoretical guarantees for PC-Diffusion:~\textbf{1)}~PC-Diffusion ensures that the preference-guided distributions are consistently propagated across timesteps.~\textbf{2)}~The training objective of the preference classifier is equivalent to DPO, but does not require a reference model.~\textbf{3)}~The proposed preference-guided correction can progressively steer generation toward preference-aligned regions.~Empirical results show that PC-Diffusion achieves comparable preference consistency to DPO while significantly reducing training costs and enabling efficient and stable preference-guided generation.
\end{abstract}

\section{Introduction}
\IEEEPARstart{d}{iffusion} models~\cite{dhariwal2021diffusion,song2020score} have demonstrated remarkable success in conditional image synthesis~\cite{rombach2022high,yang2023eliminating,xu2024sgdm,wang2025adaptive,zhao2025catversion} and video generation~\cite{wang2025uniadapter,yang2025spatio,chen2025videodreamer}, producing high-resolution and photorealistic outputs conditioned on diverse inputs, such as text prompts, sketches, and structural representations. These capabilities have enabled broad applications across domains, including digital art~\cite{liu2025ctigen,wang2024diffusion,ahn2024dreamstyler,chen2025pimg,zhang2024mmginpainting}, content creation~\cite{zhang2023adding,wang2025multi,zhang2025multi,jiang2024animediff} and other visual applications~\cite{huang2025learning,qing2024diffuie,qiu2024dbsr}. However, despite their generative prowess, diffusion models often fall short in capturing subjective aspects of human preference, such as aesthetic quality and semantic relevance, leading to a misalignment between model outputs and user intent.

To address the challenges above, researchers adapt successful preference learning techniques from large language models~\cite{christiano2017deep,ouyang2022training,song2024preference,stiennon2020learning} for diffusion-based generation. Among these, Direct Preference Optimization (DPO)~\cite{rafailov2023direct} has emerged as a promising approach by directly learning from preference data to steer models toward more desirable outputs. While DPO-like methods~\cite{meng2024simpo,rafailov2023direct,gu2024diffusion,borso2025d3po} have shown strong alignment performance in language tasks, their adaptation to diffusion models presents two critical challenges:~\textbf{1)~High computational cost},~due to the requirement of fine-tuning the entire models, leading to significant overhead.~\textbf{2)~Sensitivity to reference model quality},~as their reliance on a reference model can impact overall performance and increase training complexity.

To address these challenges, we introduce \textbf{PC-Diffusion}, a novel framework for human preference alignment in diffusion models that removes the reliance on reference models and avoids costly fine-tuning. Unlike DPO-like approaches, PC-Diffusion leverages a lightweight, trainable preference classifier~(PC) to model relative preferences between samples, thereby decoupling preference learning from the base generative model. We provide theoretical support for PC-Diffusion from three perspectives: \textbf{1)} its preference-guided transition process guarantees consistent propagation of aligned distributions across timesteps; \textbf{2)} its training objective is provably equivalent to DPO without the need for a reference model; and \textbf{3)} its preference-guided correction progressively steers generation toward human-preferred outputs.

Our contributions can be summarized as follows:
\begin{itemize}
    \item \textbf{Reference-free Preference Alignment Framework:}~We propose PC-Diffusion, a novel, reference-free preference alignment framework that significantly cuts computational costs. PC-Diffusion employs a lightweight, trainable classifier to model relative preferences, decoupling this from the generative model. This avoids large model fine-tuning, eliminates reference model dependency, and substantially reduces training expenses.

    \item \textbf{Theoretical innovation and completeness analysis:}~We provide strong theoretical validation for PC-Diffusion, demonstrating: \textbf{1)} its preference-guided transition process guarantees consistent propagation of aligned distributions across timesteps; \textbf{2)} its training objective equates to a reference-free DPO; and \textbf{3)} its preference-guided correction progressively steers generation towards human-favored regions.

    \item \textbf{Significant performance advantages:}~Extensive experiments show PC-Diffusion's significant performance advantages. On standard benchmarks, it achieves preference consistency comparable to or superior to DPO, while being markedly more efficient and stable. PC-Diffusion thus offers a more reliable, practical alignment solution for real-world generative tasks.
\end{itemize}
\section{Related Work}
This section briefly reviews the literature on preference alignment in large language models and diffusion models.

\subsection{Preference Alignment in Large Language Models}

To more effectively align Large Language Models (LLMs) with human preferences, the community has developed a multi-stage alignment paradigm. Early work primarily focused on Supervised Fine-Tuning (SFT)~\cite{chen2020adversarial,prottasha2022transfer,lee2018training,huo2025enhancing}, which provides LLMs with an initial ability to follow instructions by training on high-quality instruction–output pairs. Although SFT improves instruction-following behavior, it fails to capture nuanced human preferences, particularly in human subjective tasks. To address these limitations, Reinforcement Learning from Human Feedback (RLHF)\cite{griffith2013policy,lin2020review,shen2024improving} was introduced. RLHF frameworks typically involve collecting human preference data to train a Reward Model (RM), which then guides the LLM via reinforcement learning algorithms such as Proximal Policy Optimization (PPO). This approach has shown significant success in high-profile systems like InstructGPT\cite{ouyang2022training} and ChatGPT~\cite{dalvi2024rlhf}. However, RLHF is notoriously complex and resource-intensive, requiring extensive preference annotation, reward modeling, and policy optimization. To reduce this complexity, Direct Preference Optimization (DPO)\cite{rafailov2023direct} has been proposed as a simple and effective alternative to RLHF. By directly optimizing on preference pairs using a loss derived from the Bradley-Terry model\cite{hunter2004mm}, DPO eliminates the need for explicit reward models and reinforcement learning. Variants \cite{meng2024simpo, lu2025inpo} further improve efficiency and robustness. Owing to its simplicity and strong empirical results, DPO has become a popular alignment method in LLMs.

\subsection{Preference Alignment in Diffusion Models}

In parallel with LLMs, diffusion models have demonstrated strong generative capabilities across various modalities such as image synthesis, video generation, and 3D modeling. However, aligning their outputs with human intent remains challenging, especially regarding subjective preferences like aesthetics, style, and semantics. Although RLHF-like frameworks have been adapted to image generation~\cite{yang2024using,fan2023dpok,yuan2024instructvideo,black2023training}, they suffer from costly and unstable optimization due to ambiguous visual reward functions and the high dimensionality of pixel space. To address these issues, recent work extends Direct Preference Optimization (DPO) to diffusion models~\cite{wallace2024diffusion,zhu2025dspo,majumder2024tango,shan2025forward,cheng2025discriminator}, enabling direct supervision from image preference pairs without explicit reward models or complex RL. While effective, DPO-like methods still face two major limitations: \textbf{1) high computational cost}, due to full model fine-tuning and \textbf{2) reliance on a reference model}, which may introduce instability and bias. To overcome these issues, we propose \textbf{PC-Diffusion}, a novel framework for human preference alignment in diffusion models. Specifically, PC-Diffusion introduces a lightweight, trainable preference classifier (PC) that models relative preference between image pairs independently of the base generative model. This decoupling enables efficient preference alignment without full model updates or reliance on a reference model. 
\section{Preliminaries}
This section presents the background for our method, \textbf{PC-Diffusion}, including Denoising Diffusion Probabilistic Models (DDPM) as the generative backbone and Direct Preference Optimization (DPO) as a representative preference alignment approach.
\subsection{Denoising Diffusion Probabilistic Models (DDPM)}
\label{subsec:ddpm}
Diffusion models define a generative process as the reversal of a fixed forward noising procedure. The forward process gradually perturbs a data sample $x_0 \sim q(x_0)$ over $T$ steps via a Markov chain:
\begin{equation}
    q(x_t \mid x_{t-1}) = \mathcal{N}(x_t; \sqrt{1 - \beta_t} x_{t-1}, \beta_t \mathbf{I}),
\end{equation}
where $\{\beta_t\}_{t=1}^T$ denotes the variance schedule. In the reverse process, a neural network $\epsilon_\phi(x_t, t)$ is trained to approximate the posterior $q(x_{t-1} \mid x_t)$. The reverse transition is defined as:
\[
p_\phi(x_{t-1} \mid x_t) = \mathcal{N}(x_{t-1}; \mu_\phi(x_t, t), \sigma_t^2 \mathbf{I}),
\]
and sampling proceeds from $x_T \sim \mathcal{N}(0, \mathbf{I})$ using:
\begin{align}
x_{t-1} &= \underbrace{\frac{1}{\sqrt{\alpha_t}}\left(x_t - \frac{1 - \alpha_t}{\sqrt{1 - \bar{\alpha}_t}} \epsilon_\phi(x_t, t)\right) + \sigma_t \epsilon}_{\text{DDPM Sampling}}, \quad \epsilon \sim \mathcal{N}(0, \mathbf{I}) \nonumber \\ 
        &= \texttt{DDPM}(x_t, \epsilon_\phi(x_t, t)).
\label{eq:ddpm-sampling}
\end{align}
We follow the notation in DDPM~\cite{ho2020denoising}: $\alpha_t = 1 - \beta_t$, $\bar{\alpha}_t = \prod_{i=1}^t \alpha_i$, $\mu_\phi(x_t, t) = \frac{1}{\sqrt{\alpha_t}}\left( x_t - \frac{\beta_t}{\sqrt{1 - \bar{\alpha}_t}} \epsilon_\phi(x_t, t) \right)$, and $\sigma_t^2 = \frac{1 - \bar{\alpha}_{t-1}}{1 - \bar{\alpha}_t} \beta_t$.

\subsection{Direct Preference Optimization (DPO)}
\label{subsec:dpo}
Direct Preference Optimization (DPO)~\cite{rafailov2023direct} provides an effective approach for aligning generative models with human preferences by directly optimizing a policy $\pi_\theta(x)$ from preference pairs $(x^w, x^l)$, where $x^w$ is preferred over $x^l$. The objective maximizes the relative likelihood of preferred samples over less preferred ones:
\begin{equation}
\begin{aligned}
\mathcal{L}_{\text{DPO}}(\pi_\theta; \pi_{\text{ref}})
&= -\mathbb{E}_{(x^w, x^l)\sim\mathcal{D}} \\
& \Bigg[ \log \sigma \Big(
   \beta \log \frac{\pi_\theta(x^w)}{\pi_{\text{ref}}(x^w)}
   - \beta \log \frac{\pi_\theta(x^l)}{\pi_{\text{ref}}(x^l)}
\Big) \Bigg]
\end{aligned}
\label{dpo}
\end{equation}
where $\mathcal{D}$ denotes the dataset of human preference pairs, $\pi_{\text{ref}}(x)$ is a reference policy and $\beta$ controls the deviation from it.

However, DPO and its variants face two critical challenges:~\textbf{1)~High computational cost}. They need to fine-tune the full base model $\pi_\theta(x)$, which can be computationally expensive, and~\textbf{2)~Sensitivity to reference model quality.} The performance is sensitive to the choice of the reference model $\pi_{ref}(x)$, which may introduce bias or instability.

These limitations motivate our work on \textbf{PC-Diffusion}, which avoids modifying the base model and eliminates the need for a reference policy by introducing a lightweight preference classifier $\mathcal{S}_\theta(x)$.

\section{Preference Classifier Guidance}
\label{sub:PCG}
In this section, we introduce our preference classifier, which steers the diffusion model to generate results aligned with human preferences. First, we explain how the classifier adjusts the model's generation probabilities to better reflect human preferences, without the need to retrain the model. Next, we present the training objective that injects human preference priors into the preference classifier. Finally, we outline the procedure for human preference-guided sampling using the human preference prior embedded in the preference classifier.

\subsection{The Definition for Preference Classifier}
\label{sub:def-pcg}
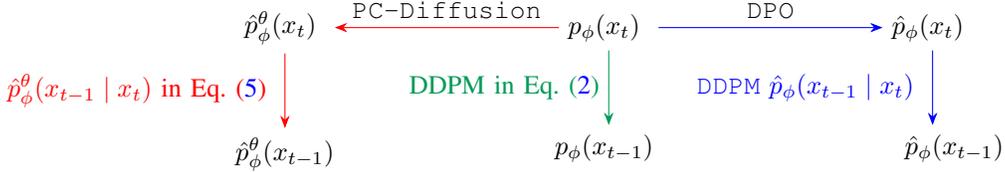
\begin{figure*}[t]
\centering
\resizebox{0.75\linewidth}{!}{%
\begin{tikzpicture}[node distance=3.0cm, every node/.style={align=center}, >=Stealth, baseline]

\node (p1) { $\hat{p}_\phi^\theta(x_t)$ };
\node (p2) [right=of p1] { $p_\phi(x_t)$ };
\node (p3) [right=of p2] { $\hat{p}_\phi(x_t)$ };

\node (p4) [below=1.0cm of p1] {$\hat{p}_\phi^\theta(x_{t-1})$};
\node (p5) [below=1.0cm of p2] { $p_\phi(x_{t-1})$ };
\node (p6) [below=1.0cm of p3] { $\qquad \hat{p}_\phi(x_{t-1})$ };

\draw[->, draw=red] 
(p1) -- node[left, align=center]{
    \textcolor{red}{$\hat{p}_\phi^\theta(x_{t-1} \mid x_t)$} \textcolor{red}{in Eq.~\eqref{eq:pc_transition}}
} (p4);
\draw[->, draw=ForestGreen] (p2) -- node[left]{\textcolor{ForestGreen}{\text{DDPM} in Eq.~\eqref{eq:ddpm-sampling}}} (p5);
\draw[->, draw=blue] (p3) -- node[left]{ \textcolor{blue}{\texttt{DDPM} $\hat{p}_\phi(x_{t-1}\mid x_t)$} } (p6);

\draw[<-, draw=red] (p1) -- node[above]{\texttt{PC-Diffusion}} (p2);
\draw[->, draw=blue] (p2) -- node[above]{\texttt{DPO}} (p3);

\end{tikzpicture}%
}
\caption{Relationship between PC-Diffusion, DPO, and DDPM.
Nodes: top row ($t$) and bottom row ($t\!-\!1$) list, from left to right, the preference-guided process $\hat{p}_\phi^\theta(\cdot)$, the standard DDPM process $p_\phi(\cdot)$, and the DPO process $\hat{p}_\phi(\cdot)$.
Horizontal arrows: method transformations at the same timestep—\textcolor{red}{PC-Diffusion} steers DDPM toward human-preferred distributions, while \textcolor{blue}{DPO} maps DDPM to the DPO process.
Vertical arrows: transitions $x_t\!\to\! x_{t-1}$ within each column: \textcolor{red}{PC-Diffusion} in Eq.~\eqref{eq:pc_transition}, \textcolor{ForestGreen}{DDPM} in Eq.~\eqref{eq:ddpm-sampling}, and \textcolor{blue}{DPO} transitions.}
\label{fig:pcg-dpo-ddpm}
\end{figure*}

In preference-misaligned generation tasks, DDPM naturally propagates the preference-misaligned distribution $p_{\phi(x_t)}^t$ to $p_{\phi(x_{t-1})}^{t-1}$ from timestep $t$ to $t-1$ through DDPM transition probability $p_\phi(x_{t-1} \mid x_t)$.~For simplicity, we omit the time step $t$.~To match human preference, recent methods such as DPO, achieve preference-aligned distribution $\hat{p}_\phi(x_t)$ from $p_\phi(x_t)$ by optimizing the entire model, thereby enabling transform $\hat{p}_\phi(x_t)$ to $\hat{p}_\phi(x_{t-1})$ through updated DDPM transition probability $\hat{p}_\phi(x_{t-1} \mid x_t)$.~The overall process is illustrated in the Fig.~\ref{fig:pcg-dpo-ddpm}, with \textcolor{ForestGreen}{green arrows} representing the DDPM transition and \textcolor{blue}{blue arrows} representing the DPO method.

Different from DPO-like methods, we propose \textbf{PC-Diffusion}, which transforms $p_\phi(x_t)$ into preference-guided distribution $\hat{p}_\phi^\theta(x_t)$ via a \textbf{preference classifier }$\mathcal{S}_\theta(x)$, and propagates $\hat{p}_\phi^\theta(x_t)$ to $\hat{p}_\phi^\theta(x_{t-1})$ through the preference-guided transition probability $\hat{p}_\phi^\theta(x_{t-1} \mid x_t)$.~This process is depicted by the \textcolor{red}{red arrows} in the figure and the components of this process are defined as follows.

\begin{definition}[Preference Classifier Guidance] 
\label{def:preference-classifier-and-induced-distributions}
The trainable \textbf{preference classifier} $\mathcal{S}_\theta(x): \mathbb{R}^d \to (0,1)$ assigns a human preference score to sample $x$~(higher scores indicate greater preference).~Based on $\mathcal{S}_\theta(x)$, we define the \textbf{preference-guided distribution} $\hat{p}_\phi^\theta(x_t)$ and the corresponding \textbf{preference-guided transition} $\hat{p}_\phi^\theta(x_{t-1} \mid x_t)$ respectively:
\begin{align}
    &\hat{p}_\phi^\theta(x_t) = N_t p_\phi(x_t) \exp\left( \log \mathcal{S}_\theta(x_t) \right), 
    \label{eq:pc_distribution_old} \\
    &\hat{p}_\phi^\theta(x_{t-1} \mid x_t) = p_\phi(x_{t-1} \mid x_t) 
    \exp\Big( \log \mathcal{S}_\theta(x_{t-1}) \nonumber \\
    &\hspace{6em} - \log \mathcal{S}_\theta(x_t) \Big),
    \label{eq:pc_transition}
\end{align}
where $N_t$ is a normalization term.
\end{definition}

Since DPO-like methods transform preference-aligned distribution $\hat{p}_\phi(x_t)$ to $\hat{p}_\phi(x_{t-1})$ through training the entire model, it is necessary for us to investigate whether our proposed formulation in Eq.~\eqref{eq:pc_transition} can propagate \(\hat{p}_\phi^\theta(x_{t})\) to \(\hat{p}_\phi^\theta(x_{t-1})\).Theorem~\ref{thm:preference-classifier-transition} guarantees the correctness of the preference-guided distribution transformation.
\begin{theorem}
\label{thm:preference-classifier-transition}
(Proof in the supplementary material)~
Given a preference classifier \( \mathcal{S}_\theta(x) \), suppose that \( x_t \) is sampled from $\hat{p}_\phi^\theta(x_t)$, and then apply the transition $ \hat{p}_\phi^\theta(x_{t-1} \mid x_t)$, we can generate:
 \begin{equation}
        \hat{p}_\phi^\theta(x_{t-1}) = N_{t-1}p_\phi(x_{t-1}) \exp\left( \log \mathcal{S}_\theta(x_{t-1}) \right),
        \label{eq:pc_distribution}
    \end{equation}
    where \( N_{t-1} \) is normalization term.
\end{theorem}

Thus, this theorem guarantees that we can achieve the preference-guided distribution by optimizing only the preference classifier $\mathcal{S}_\theta(x)$ without altering the preference-misaligned distribution $p_\phi(x_t)$.

\subsection{The Training Objective for Preference Classifier}
\label{sub:PCG-Training}
Having established that the preference classifier $\mathcal{S}_\theta(x)$ governs the transformation from preference-misaligned $p_\phi(x_t)$ to preference-guided distributions $\hat{p}_\phi^\theta(x_t)$, we now describe how to train $\mathcal{S}_\theta(x)$ to properly reflect human preferences.~Unlike traditional methods such as DPO, which optimize the entire model to match preference distributions, our PC-Diffusion focuses solely on learning the preference through $\mathcal{S}_\theta(x)$.

Specifically, by rewriting the decomposition in Eq.~\eqref{eq:pc_transition}, we observe that the adjustment toward preference alignment depends solely on the relative preference scores provided by $\mathcal{S}_\theta(x)$, leading to the ratio: $ \frac{\hat{p}_\phi^\theta\left(x_{t-1} \mid x_t\right)}{p_\phi\left(x_{t-1} \mid x_t\right)}=\frac{\mathcal{S}_\theta\left(x_{t-1}\right)}{\mathcal{S}_\theta\left(x_t\right)}$.

Crucially, this insight implies that we only need to ensure $\mathcal{S}_\theta(x)$ accurately reflects human preference. This allows us to derive a reference-free training objective called PC-Diffusion Loss for $\mathcal{S}_\theta(x)$ using standard preference tuple $\left(x^w,x^l\right)\sim \mathcal{D}$, as formalized in Eq.~\eqref{eq:PCG-loss}:
\begin{equation}
\begin{aligned}
&\mathcal{L}_{\mathrm{PC\text{-}Diffusion}} \!\left(\theta\right)
= -\mathbb{E}_{\substack{
  (x_t^w, x_{t-1}^w), 
  (x_t^l, x_{t-1}^l) \sim \mathcal{D}, t \sim \mathcal{U}(0,T)
}} \\
& \qquad \qquad \Bigg[
   \log \sigma\Big(
      \beta T \log \tfrac{\mathcal{S}_\theta(x_{t-1}^w)}{\mathcal{S}_\theta(x_t^w)}
      - \beta T \log \tfrac{\mathcal{S}_\theta(x_{t-1}^l)}{\mathcal{S}_\theta(x_t^l)}
   \Big)
\Bigg],
\end{aligned}
\label{eq:PCG-loss}
\end{equation}
where \(\sigma(\cdot)\) denotes the sigmoid function, $\mathcal{D}$ is the preference dataset, and $T$ is the total number of diffusion steps. This objective is guaranteed by Theorem~\ref{thm:preference-dpo-connection}.
\begin{theorem}
\label{thm:preference-dpo-connection}
(Proof in the supplementary material)~
Given $\mathcal{S}_\theta(x)$, $\hat{p}_\phi^\theta(x_{t-1} \mid x_t)$, and human preference samples in different timesteps, the PC-Diffusion training objective in Eq.~\eqref{eq:PCG-loss} can be derived by rewriting the DPO loss~as defined in Eq.~\eqref{dpo} without requiring a reference model. 
\end{theorem}

Theorem~\ref{thm:preference-dpo-connection} eliminates the need for a reference model, which is required in DPO-like methods as described in Eq.~\eqref{dpo}.~This is because the relative preference between samples is captured entirely by the outputs of the preference classifier $\mathcal{S}_\theta$, without relying on a reference model.

The design of PC-Diffusion offers several practical advantages:~\textbf{1) Reference-free Optimization:}~The optimization no longer requires a reference model, simplifying training and crucially avoiding reference model instabilities or biases;~\textbf{2) High Efficiency:}~Training only the lightweight classifier $\mathcal{S}_\theta$ enables plug-and-play integration with pre-trained diffusion models, eliminating costly base model fine-tuning.

\subsection{Human Preference-Guided Sampling}
\label{sub:PCG-Sampling}
Diffusion models require sampling from the reverse transition distribution \( p_\phi(x_{t-1} \mid x_t) \) at each timestep \( t \). DDPM provides a standard approach by modeling this distribution as a Gaussian~\eqref{eq:ddpm-gaussian} and computing samples via a noise prediction network $\epsilon_\phi$~\eqref{eq:ddpm-solve}.~The corresponding formulations are presented below:
\begin{align}
    p_\phi(x_{t-1} \mid x_t) &= \mathcal{N}(x_{t-1}; \mu_\phi(x_t), \sigma_t^2 \mathbf{I}),
    \label{eq:ddpm-gaussian}\\
    x_{t-1} &= \text{DDPM}(x_t, \epsilon_\phi(x_t, t)),
    \label{eq:ddpm-solve} 
\end{align}
where \( \mu_\phi(x_t) \) and \( \sigma_t \) are determined by the learned noise prediction network $\epsilon_\phi$.

We modify $p_\phi(x_{t-1} \mid x_t)$ into a preference-guided transition~$\hat{p}_\phi^\theta(x_{t-1} \mid x_t)$~\eqref{eq:PCG-Gaussian} as described in~\ref{sub:def-pcg} and adopt an approximate sampling strategy \eqref{eq:denosing} to sample $x_{t-1}$~from Eq.~\eqref{eq:PCG-Gaussian}.~Specifically, we summarize the preference-guided formulation as follows:
\begin{align}
\hat{p}_\phi^\theta(x_{t-1} \mid x_t) 
&= \mathcal{N}\!\left(
      x_{t-1};\, \mu_\phi(x_t),\, \sigma_t^2 \mathbf{I}
    \right) \nonumber\\
&\quad 
    \exp\!\Big(
      \log \mathcal{S}_\theta(x_{t-1})
      - \log \mathcal{S}_\theta(x_t)
    \Big),
\label{eq:PCG-Gaussian}
\\[0.5em]
x_{t-1} 
&\approx 
\underbrace{\text{DDPM}\!\left(
      x_t,\, \epsilon_\phi(x_t,t)
    \right)}_{\text{DDPM Sampling}}
\nonumber\\
&\quad+\underbrace{
    \gamma \sigma_t^2\, \nabla_{x_t} \log \mathcal{S}_\theta(x_t)
}_{\text{Preference Correction}},
\label{eq:denosing}
\end{align}
where \( \gamma \) is a tunable guidance weight.~Theorem~\ref{thm:pcg-sampling} guarantees that Eq.~\eqref{eq:denosing} serves as a good approximate sampling strategy for drawing samples from the distribution defined in Eq.~\eqref{eq:PCG-Gaussian}.

\begin{theorem}
\label{thm:pcg-sampling}
(Proof in the supplementary material)~
Given the $\hat{p}_\phi^\theta(x_{t-1} \mid x_t)$~in Eq.~\eqref{eq:PCG-Gaussian}, and  $x_t$ is sampled from $ x_t \sim \hat{p}_\phi^\theta(x_t)$ , the $x_{t-1}$ can be approximately sampled by Eq.~\eqref{eq:denosing}.
\end{theorem}
The sampling strategy in Eq.~\eqref{eq:denosing} is an approximate method and therefore does not strictly guarantee a monotonic improvement in the preference score across timesteps~ (\textit{e.g.},~ $\mathcal{S}_\theta(x_{t-1}) \geq \mathcal{S}_\theta(x_t)$). To address this, we incorporate a rejection sampling mechanism to enforce monotonicity.

\textbf{Rejection Sampling.}~We adopt a rejection sampling strategy to enforce the monotonicity constraint $\mathcal{S}_\theta(x_{t-1}) \geq \mathcal{S}_\theta(x_t)$ across timesteps, ensuring progressive alignment with human preferences. The procedure is formally defined as in Eq.~\eqref{eq:reject}:
\begin{equation}
    x_{t-1} =
    \begin{cases}
    \tilde{x}_{t-1}, & \text{if } \mathcal{S}_\theta(\tilde{x}_{t-1}) \geq \mathcal{S}_\theta(x_t), \\
    \text{Z-sampling}, & \text{otherwise}.
    \end{cases}
    \label{eq:reject}
\end{equation}
where $\tilde{x}_{t-1}$ is the candidate sample following Eq.~\eqref{eq:denosing}.~However, if no valid candidate is found within $M$ attempts (default $5$), rejection sampling becomes inefficient.~To address this issue, we adopt Z-sampling~\cite{bai2024zigzag}, which enriches $x_t$ with additional semantic information by applying an inversion function~$\Phi$, to decrease probability of reject sampling.~The updated \( x_t \) is obtained via:
\begin{equation}
    x_t = \Phi(\tilde{x}_{t-1}, t)
    \label{eq:inversion}
\end{equation}

\begin{algorithm}[!t]
\caption{Constrained Preference-Guided Sampling with Z-Sampling}
\label{alg:constrained_pc}
\begin{algorithmic}[1]
\State \textbf{Input:} Initial noise: \( x_T \sim \mathcal{N}(0, \mathbf{I}) \); preference classifier: \( \mathcal{S}_\theta(x) \); Denoising at timestep $t$: \( \epsilon_\phi(x_t, t) \), Inversion at timestep $t$: $\Phi(x_t,t)$, variance schedule: \( \{ \sigma_t^2 \}_{t=1}^T \); guidance weight: \( \gamma \); maximum resampling attempts: \( M \).
\State \textbf{Output:} Clean image $x_0$
\For{$t = T, T{-}1, \ldots, 1$}
    \State \# \textbf{Candidate Generation}
    \State \(  \tilde{x}_{t-1} = DDPM(x_t, \epsilon_\phi(x_t, t)) + \gamma \sigma_t^2 \nabla_{x_t} \log \mathcal{S}_\theta(x_t)  \) \hfill \(\triangleright\) Sample follows Eq.~\eqref{eq:denosing}
    \State Initialize resampling counter: \( m \leftarrow 0 \)
    \While{ \( \mathcal{S}_\theta(\tilde{x}_{t-1}) < \mathcal{S}_\theta(x_t) \) \textbf{and} \( m < M \)}
        \State \quad \# \textbf{Resample candidate using Z-Sampling}
        \State \quad \(x_t=\Phi(\tilde{x}_{t-1},t)  \) \hfill \(\triangleright\) Inverse follows Eq.~\eqref{eq:inversion}
        \State \quad \( \tilde{x}_{t-1} = DDPM(x_t, \epsilon_\phi(x_t,t)) + \gamma \sigma_t^2 \nabla_{x_t} \log \mathcal{S}_\theta(x_t) \)
        \State \quad Update counter: \( m \leftarrow m + 1 \)
    \EndWhile
    \State  \text{Accept } $\tilde{x}_{t-1}$ \hfill \(\triangleright\) Rejection sampling follows Eq.~\eqref{eq:reject}
\EndFor
\State \textbf{return} \( x_0 \)
\end{algorithmic}
\end{algorithm}

\begin{table*}[!t] 
\caption{Quantitative Win-rate Comparison Using Automated Preference Metrics. We evaluate the alignment performance of diffusion models using prompts from HPS and PartiPrompts across various evaluators. SD1.5 serves as base models. Win rates above $50\%$—indicating superior performance over the baseline—are highlighted in bold. }
\label{tab:quantitative_cmp} 

\vskip 0.04in 

\centering 

\resizebox{1\linewidth}{!}{
\begin{tabular}{>{\raggedright\arraybackslash}p{2.55cm} 
                >{\centering\arraybackslash}p{1.2cm}
                >{\centering\arraybackslash}p{1.2cm}
                >{\centering\arraybackslash}p{1.2cm}
                >{\centering\arraybackslash}p{1.2cm} 
                >{\centering\arraybackslash}p{1.2cm}
                >{\centering\arraybackslash}p{1.2cm}
                >{\centering\arraybackslash}p{1.2cm}
                >{\centering\arraybackslash}p{1.2cm}}  

\toprule
& \multicolumn{4}{c}{\texttt{Partiprompt}} 
& \multicolumn{4}{c}{HPS benchmarks} \\ 
\cmidrule(lr){2-5} \cmidrule(lr){6-9}

& PickScore & HPS &Aesthetics &  CLIP & PickScore & HPS & Aesthetics & CLIP \\
\midrule

vs. SD1.5   & \textbf{61.55} & \textbf{83.43} & \textbf{71.25} & \textbf{58.63} & \textbf{75.24}  & \textbf{82.63} & \textbf{65.85} & \textbf{58.45} \\
vs. SD1.5-DPO   & \textbf{56.63} & \textbf{71.25} & \textbf{65.11} & 43.27 & \textbf{51.25} & \textbf{68.53} & \textbf{55.29}  & \textbf{54.36}  \\
vs. SD1.5-SPO   & \textbf{50.03} & \textbf{63.29} & 45.36 & \textbf{61.13} &  43.26 & \textbf{55.01} & 36.24  & \textbf{72.28}  \\
vs. SD1.5-KTO   & \textbf{56.63} & 43.22 & \textbf{51.68} & 45.31 & \textbf{52.35} & 40.67 & \textbf{51.73} & \textbf{51.85} \\
\bottomrule

   \label{fig:t2i_comp}
\end{tabular}
}
\vspace{-0.75cm}
\end{table*}

\section{Experiments}
This section presents a comprehensive empirical evaluation of our proposed \textbf{PC-Diffusion} framework for aligning diffusion model outputs with human preferences. We first detail our common experimental setup (Section~\ref{subsec:exp}). We then conduct thorough evaluations of PC-Diffusion in three representative alignment scenarios: \textbf{1) Aesthetic Alignment} (Section~\ref{subsec:aesthetic}), which assesses the ability to enhance perceptual quality and visual appeal; \textbf{2) Text-to-Image Alignment} (Section~\ref{subsec:t2i}), which examines semantic consistency between generated images and textual prompts; and \textbf{3) Conditional Alignment} (Section~\ref{subsec:conditional}), which evaluates the ability to align generated images with additional conditioning inputs.

\begin{figure*}[!t]
    \centering
    \includegraphics[width=0.85\textwidth]{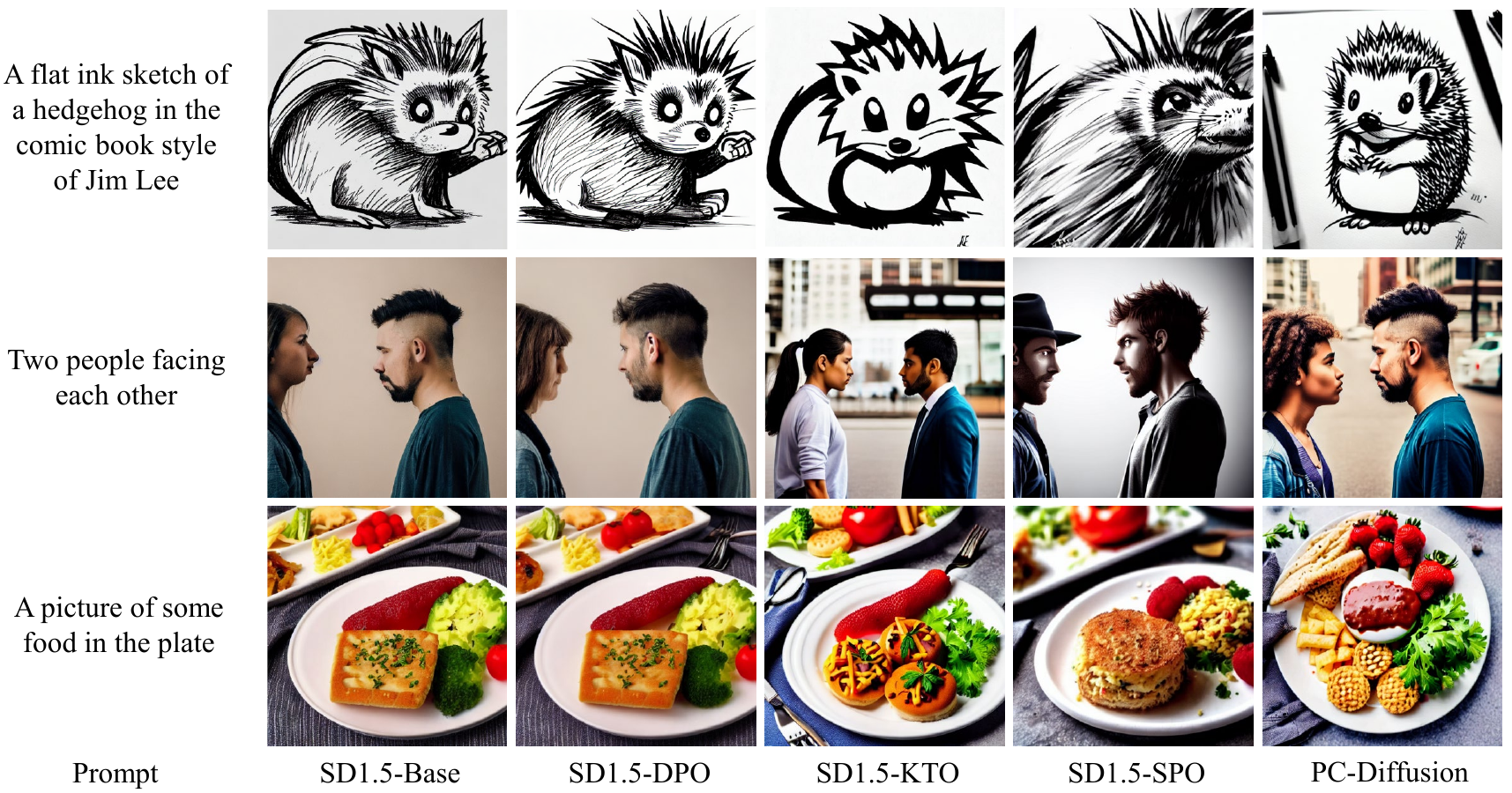} 
    \caption{Qualitative Comparison for Aesthetic Alignment. We present an alternative alignment approach based on PC-Diffusion, which steers pretrained diffusion models toward human-preferred outputs via a preference-guided correction term, without fine-tuning the entire model. Applied to SD1.5~\cite{rombach2022high}, PC-Diffusion achieves superior visual and textual alignment compared to baselines including SD1.5-Base, SD1.5-DPO~\cite{wallace2024diffusion}, SD1.5-KTO~\cite{li2024aligning}, SD1.5-SPO~\cite{liang2024step}.}
    \vspace{-0.75cm}
\label{fig:aligned_grid}
\end{figure*}

\subsection{Experimental Setup}
\label{subsec:exp}
\textbf{Models and Datasets:~}We use a frozen Stable Diffusion v1.5 (SD1.5)~\cite{rombach2022high} as the foundational base generative model throughout all our experiments. For the aesthetic and text-image alignment tasks, our preference classifier $\mathcal{S}_\theta(x)$ is trained on the Pick-a-Pic dataset~\cite{kirstain2023pick} and evaluated on HPSv2~\cite{wu2023human} and Parti-Prompts collection~\cite{yu2022scaling}. For the conditional alignment task, which involves $9$ conditions (Depth, HED, Canny, Segmentation, Normal, Sketch, MLSD, Box, and OpenPose) derived from the COCO-2017 dataset~\cite{lin2014microsoft} (conditions obtained following methodologies in~\cite{zhao2023uni,qin2023unicontrol}), its corresponding preference data is simply constructed. This involves labeling original COCO images paired with their accurately matching conditions as preferred, and those paired with deliberately mismatched conditions as non-preferred.

\textbf{Training Setup.}~Our preference classifier $\mathcal{S}_\theta(x)$ can be easily adapted to various alignment tasks by adjusting its input modalities and architecture. We implement task-specific variants as follows: 1)~For aesthetic alignment, it uses a convolutional encoder to predict a scalar score reflecting visual appeal. 2)~For text-image alignment, $\mathcal{S}_\theta(x,c)$ encodes images $x$ via a VAE~\cite{kingma2013auto} and text prompts $c$ via CLIP~\cite{radford2021learning}, combining them with cross-attention to assess semantic consistency. 3)~For the conditional alignment task, $\mathcal{S}_\theta(x, y)$ receives an image $x$ and its corresponding condition $y$. Both $x$ and $y$ are encoded into latent space using a shared frozen VAE. The classifier then computes a compatibility score between the image and its conditioning input. All classifiers are trained using the PC-Diffusion loss in Eq.~\eqref{eq:PCG-loss} for 2,000 steps with AdamW optimizer~\cite{loshchilov2017decoupled}, a batch size of 2, a learning rate of $1 \times 10^{-5}$, and a constant warm-up schedule on $8$ NVIDIA A6000 GPUs.

\textbf{Inference Setup.}~For all inference phases across the models, we employed the following configurations to ensure fair comparisons.~At each timestep $t$, the sampling update is augmented with a preference-guided correction term derived from the preference classifier $\mathcal{S}_\theta(x)$, as defined in Eq.~\eqref{eq:denosing}. We uniformly used the DDPM~\cite{dhariwal2021diffusion} with $50$ inference steps to generate images at a resolution of $512 \times512$ from the SD1.5 model.~We applied Classifier-Free Guidance (CFG)~\cite{ho2022classifier} with a guidance scale set to $7.5$.~The PC-Diffusion guidance weight hyperparameter $\gamma$ is uniformly set to $25,000$. When employing the constrained sampling mechanism detailed in Algorithm~\ref{alg:constrained_pc}, the maximum number of resampling attempts $M$ is set to $5$, and the inverse function in Z-Sampling is DDIMInverseScheduler~\cite{song2020denoising}.

\begin{figure*}[!t]
    \centering
    \includegraphics[width=0.80\textwidth]{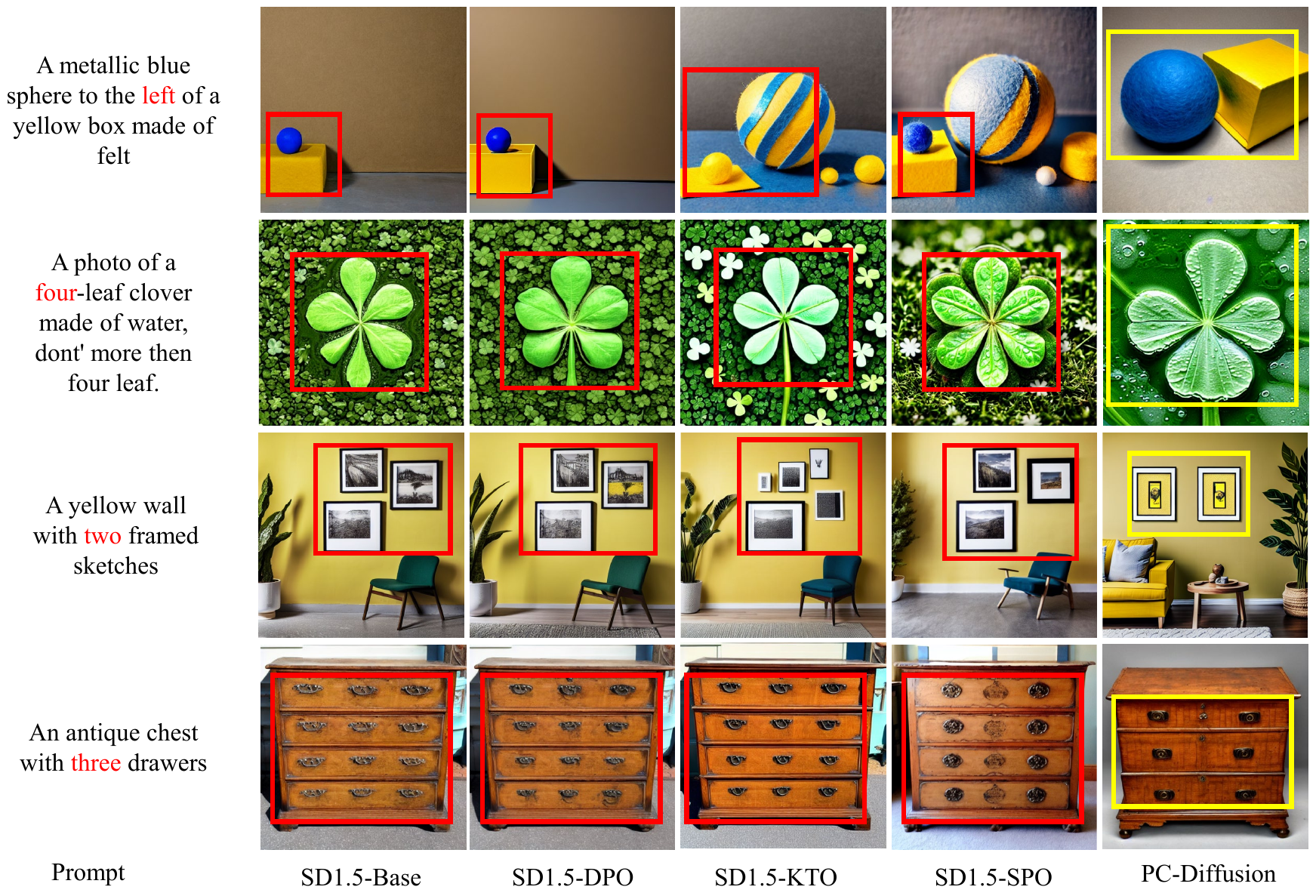} 
    \caption{Qualitative Comparison of Text-Image Alignment.~This figure showcases images generated by PC-Diffusion and other methods~(\textit{e.g., }SD1.5-DPO, SD1.5-KTO, SD1.5-SPO) for the text prompts listed on the left.~PC-Diffusion achieves enhanced semantic fidelity, particularly in accurately elements specified in the prompt, such as color and object positioning.~\textbf{Yellow boxes:} Regions matched to prompts.~\textbf{Red boxes:} Regions mismatched to prompts.} 
    \vspace{-0.4cm}
\label{fig:t2i-aligned_grid}
\end{figure*}

\begin{figure*}[!t]
\centering
  \includegraphics[width=0.86\linewidth]{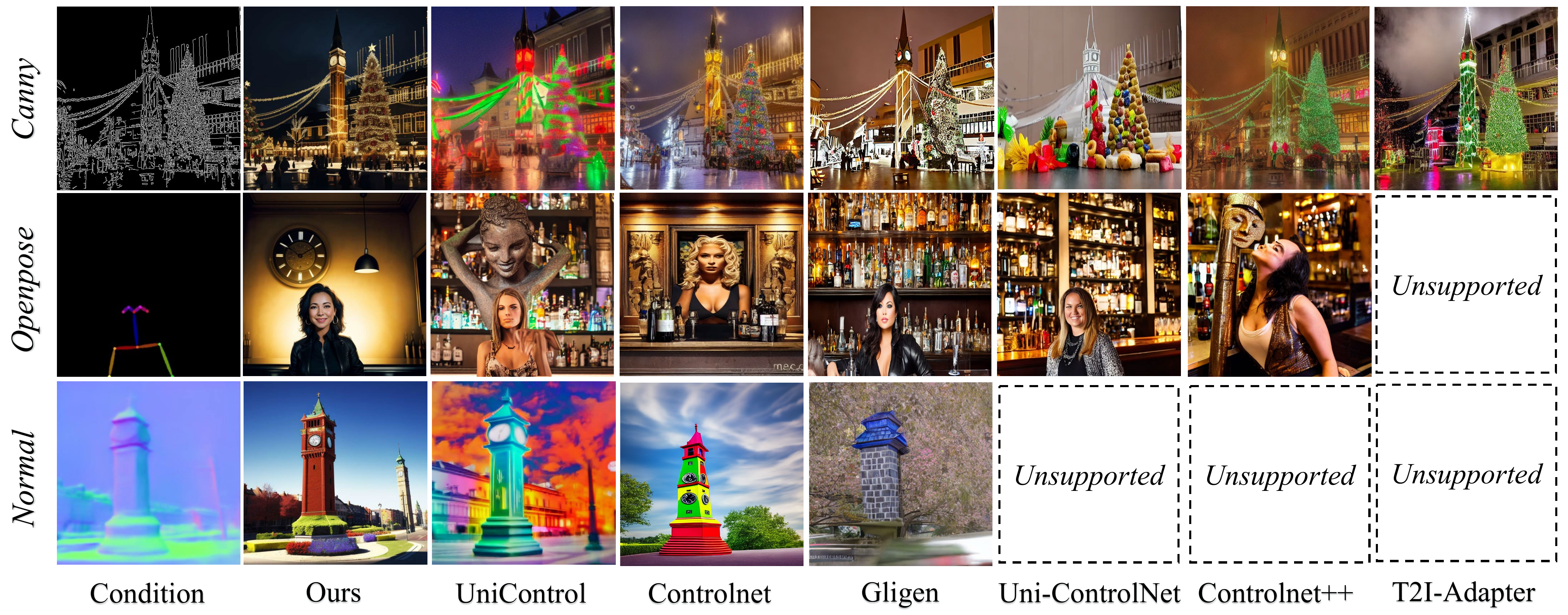}
  \caption{Comparison of existing controllable diffusion models on different conditions.~This figure illustrates a qualitative comparison of our proposed PC-Diffusion against other models, including ControlNet~\cite{zhang2023adding}, Gligen~\cite{li2023gligen}, T2I-Adapter~\cite{mou2024t2i}, UniControl~\cite{qin2023unicontrol}, Uni-ControlNet~\cite{zhao2023uni}, and ControlNet++~\cite{li2024controlnet++}.~Notably, PC-Diffusion more accurately follows the input conditions compared to baseline methods.~\textbf{Unsupported:} The method does not provide a model for image generation.} 
  \vspace{-0.5cm}
  \label{fig-comp_single}
\end{figure*}

\subsection{Aesthetic Alignment}
\label{subsec:aesthetic}
Aesthetic alignment guides generative models toward visually appealing outputs reflecting human preferences.~This subsection evaluates the efficacy of PC-Diffusion in enhancing the aesthetic quality of conditional images. To this end, we present quantitative evaluation using four automatic scores~(PickScore, HPS, Aesthetics, and CLIP), and qualitative comparisons against baseline methods.

\textbf{Qualitative Comparison.}~We present a qualitative comparison of SD1.5-PC with four baselines:~SD1.5-Base, SD1.5-DPO~\cite{wallace2024diffusion}, SD1.5-SPO~\cite{liang2024step}, and SD1.5-KTO~\cite{li2024aligning} in Fig.~\ref{fig:aligned_grid}, where SD1.5-Base denotes the original unaligned SD1.5 model. As shown in Figure~\ref{fig:aligned_grid}, PC-Diffusion produces outputs with consistently improved aesthetic alignment. By leveraging gradient-based corrections from a learned preference classifier, our method effectively integrates human preferences without model fine-tuning.~The generated images exhibit vivid color palettes, dramatic lighting, coherent compositions, fine-grained textures, and semantically faithful text-image pairs.~We also provide the results based on SDXL in the supplement material.

\textbf{Quantitative Evaluation.}~We compare our method against existing baselines, including vanilla SD, and their DPO, SPO, and KTO variants, using automated preference metrics. For automated evaluation, we assess Pick Score~\cite{kirstain2023pick}, HPS~\cite{wu2023human}, LAION Aesthetics~\cite{laionaes}, and CLIP~\cite{radford2021learning}, using prompts from the HPS benchmark~\cite{wu2023human} and PartiPrompts~\cite{yu2022scaling}. We report win rates between our method and each baseline under these metrics in Table~\ref{tab:quantitative_cmp} reports the automated evaluation results. 

\begin{table*}[!t]
\caption{Comparison of FID, CLIP text-image score with state-of-the-art methods under different conditional controls. \textbf{The best results are highlighted in bold}. '-' indicates methods that do not provide public models for testing.}
\small
\centering
\setlength{\tabcolsep}{1.8pt} 
\begin{tabular}{ccccccccccc}
\toprule
{}&{\textbf{Method}} & \textbf{Depth} & \textbf{Canny} & \textbf{Segmentation} & \textbf{HED} & \textbf{Skeleton} & \textbf{Normal}& \textbf{MLSD} & \textbf{Box} & \textbf{Sketch}\\
\midrule
 \multirow{7}{*}{\rotatebox{90}{\textit{FID}$\downarrow$}} &ControlNet     & 19.656     &\textbf{16.857} & 21.892 & 20.418     & 56.479 & 26.631 & \textbf{20.173}     &-        & -  \\
&Gligen         & 23.479     & 25.249 & 27.724 & 26.858     & 52.553 & 27.445 & -           & \textbf{24.113}  & -  \\
&T2I-Adapter    & 24.062     & 17.452 &\textbf{21.562} & -           & \textbf{36.020} & -       & -           & -       & 29.242     \\
&UniControl     & 23.981     & 19.461 & 29.241 & 19.973     & 40.659 & 29.884 & -           & 29.921 & -  \\
&Uni-ControlNet & 24.807     & 18.256 & 22.604 & 18.588     & 64.657 & -       & 27.554     & -       & 24.295   \\
&ControlNet++   &\textbf{17.408}    & 20.095 & 24.543 & 16.271     & -       & -       & -           & -        & -  \\
&\textbf{Ours}  & 22.302     & 22.431 & 22.341 & \textbf{16.264}     & 44.078 & \textbf{26.554} & 27.412     & 34.953 & \textbf{24.272}  \\
\bottomrule
\multirow{7}{*}{\rotatebox{90}{\parbox{2cm}{\centering {\textit{CLIP}\\ \textit{text-image score}}$\uparrow$}}}&ControlNet     & 0.292     & 0.311 & 0.291 & 0.295     & 0.279 & 0.283 & \textbf{0.307}    & -      & -  \\
&Gligen         & 0.308     & 0.303 & 0.294 & 0.289     & 0.269 & 0.276 & -         & 0.283 & -  \\
&T2I-Adapter    & 0.312     & 0.305 & 0.309 & -          & \textbf{0.322} & -      & -         & -      & 0.287     \\
&UniControl     & 0.317     & 0.312 & \textbf{0.326} & 0.313     & 0.317 & 0.304 & -         & \textbf{0.305} & -  \\
&Uni-ControlNet & 0.312     & 0.323 & 0.317 & 0.326     & 0.301 & -      & 0.296    & -      & \textbf{0.303}    \\
&ControlNet++   & 0.323     & 0.319 & 0.303 & 0.308     & -      & -      & -         & -      & -  \\
&\textbf{Ours}  &\textbf{0.327} & \textbf{0.325} & 0.289 & \textbf{0.327}    &0.299  & \textbf{0.306} & 0.301    & 0.281 & 0.285  \\
\bottomrule
\end{tabular}
\vspace{-0.5cm}
\label{tab:comparison_metric}
\end{table*}

\subsection{Text-to-Image Alignment}
\label{subsec:t2i}
Text-to-Image Alignment measures how accurately and comprehensively a generated image visually represents the semantic meaning and specific details of its input text prompt.
As visually demonstrated in Fig.~\ref{fig:t2i-aligned_grid}, PC-Diffusion excels at translating challenging prompts into semantically coherent and faithful images.~For instance, in the first row in Fig.~\ref{fig:t2i-aligned_grid},~with the prompt~\textit{`A metallic blue sphere to the left of a yellow box made of felt'}, PC-Diffusion accurately depicts its specified colors, materials, and precise spatial relationship, outperforming models that may falter on such combined details.~Apart from this, PC-Diffusion also correctly interprets complex object counts and attribute bindings, often achieving better outputs than other models.~These examples compellingly highlight PC-Diffusion’s enhanced capacity for text-to-image translation, significantly improving overall semantic fidelity.


\subsection{Conditional Alignment}
\label{subsec:conditional}
Conditional Alignment ensures the outputs of generative models faithfully adhere to diverse and specific input conditions~(such as Canny, Openpose, and Normal), or other guiding attributes.~ In this section, we analyze PC-Diffusion's performance through both qualitative comparisons of visual outputs and quantitative evaluations using appropriate metrics.

\textbf{Qualitative Comparison.}~We provide a qualitative comparison of the single condition in Fig.~\ref{fig-comp_single}.~We observe that while our PC-Diffusion method and other methods, including ControlNet++~\cite{li2024controlnet++}, UniControl~\cite{qin2023unicontrol}, Uni-ControlNet~\cite{zhao2023uni}, Gligen~\cite{li2023gligen}, ControlNet~\cite{zhang2023adding}, and T2I-Adapter~\cite{mou2024t2i}, all generally demonstrate competent performance in these settings, PC-Diffusion often exhibits slightly better alignment with the input conditions.~For single conditions, GLIGEN does not consider
the sketch condition, and T2I-adapter does not provide a model for HED-based generation.~Notably, PC-Diffusion achieves this enhanced fidelity by training only its lightweight preference classifier, in contrast to approaches that would require extensive fine-tuning of the larger base generative model.

\textbf{Quantitative Evaluation.}~For quantitative evaluation, we use the validation set of COCO2017 at a resolution of $512 \times 512$.~To assess generation performance, we report Fréchet Inception Distance (FID)~\cite{heusel2017gans} for perceptual quality and CLIP text-image similarity score, with detailed results presented in Table~\ref{tab:comparison_metric}.~Since some models do not provide test models for specific conditions (such as T2-apapter does not consider MLSD and HED conditions), we use '-' to indicate.~Our findings indicate PC-Diffusion effectively achieves a balance between image quality and semantic alignment with prompts.~We also report a CLIP-based aesthetic score for alignment efficacy and employ quantitative metrics to assess controllability, including SSIM for Canny, HED, MLSD, and Sketch; mAP for Skeleton and Box; MSE for Depth and Normal; and mIoU for segmentation maps. (Details in supplementary material).

\section{Conclusion}
\label{sub:conclusion}
Aligning diffusion model outputs with human preferences is essential, but methods like Direct Preference Optimization (DPO) incur high computational costs and rely on reference models. We propose \textbf{PC-Diffusion}, which uses a lightweight, trainable preference classifier (PC) to model relative preferences, decoupling preference learning from the generative model. This eliminates full-model fine-tuning and reference model dependency, reducing training overhead. Theoretically, PC-Diffusion ensures consistent propagation of preference-guided distributions, is equivalent to a reference-free DPO, and progressively steers generation toward preferred regions. Empirically, PC-Diffusion matches DPO’s alignment quality with much lower cost and stable preference-guided generation, though its performance may still depend on the quality and diversity of preference data.
\bibliographystyle{IEEEtran}
\bibliography{main}

\end{document}